\documentclass[11pt,a4paper]{article}
\usepackage[hyperref]{acl2020}
\usepackage{times}
\usepackage{latexsym}
\usepackage{graphicx}

\usepackage{microtype}

\aclfinalcopy 

\title{Customer Sentiment Analysis using Weak Supervision for Customer-Agent Chat}

\author{Navdeep Jain \\
  Comcast Applied AI Research \\
  \texttt{navdeep\_jain@comcast.com} \\}

\date{}

\begin{document}
\maketitle
\begin{abstract}
Prior work on sentiment analysis using weak supervision primarily focuses on different reviews such as movies (IMDB), restaurants (Yelp), products (Amazon).~One under-explored field in this regard is customer chat data for a customer-agent chat in customer support due to the lack of availability of free public data. Here, we perform sentiment analysis on customer chat using weak supervision on our in-house dataset. We fine-tune the pre-trained language model (LM) RoBERTa as a sentiment classifier using weak supervision.  Our contribution is as follows:1) We show that by using weak sentiment classifiers along with domain-specific lexicon-based rules as Labeling Functions (LF), we can train a fairly accurate customer chat sentiment classifier using weak supervision. 2) We compare the performance of our custom-trained model with off-the-shelf google cloud NLP API for sentiment analysis. We show that by injecting domain-specific knowledge using LFs, even with weak supervision, we can train a model to handle some domain-specific use cases better than off-the-shelf google cloud NLP API.  3) We also present an analysis of how customer sentiment in a chat relates to problem resolution.
\end{abstract}

\section{Introduction}
\begin{figure*}[t]
  \centering
  \includegraphics[width=\textwidth]{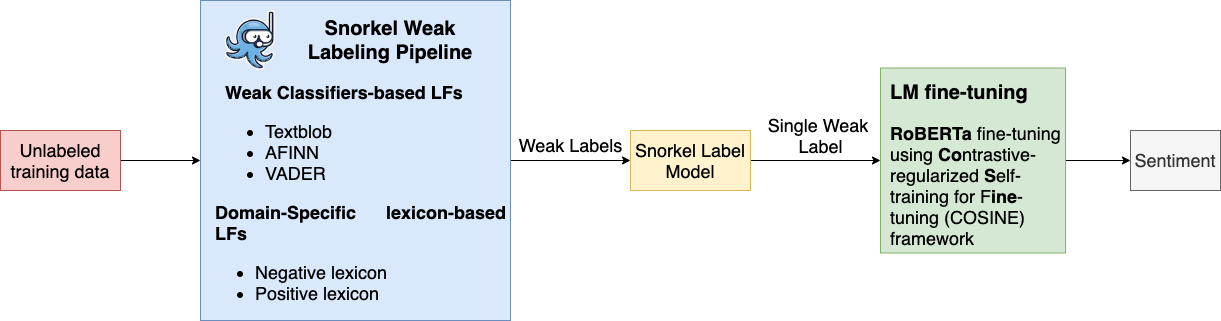}
  \caption{Our Training Framework. We apply five different labeling functions (LFs) to unlabeled training data using a snorkel pipeline. We derive a final weak label using the Snorkle label model. We fine-tune the language model (LM) using final weak labels. We use RoBERTa as LM and use Contrastive-regularized Self-training for LM fine-tuning.}
  \label{fig:sentiment_classification}
\end{figure*}
\label{sec:Introduction}
Support through chat to customers is a big part of customer support for any product/service provider. Customers must have good experiences during their chat sessions. A robust way to verify customers' experience of a chat is through customer surveys and getting Net Promoter Score (NPS). However, it's not necessary that every service provider would have such a feedback mechanism in place and even a provider has such a mechanism it's very unlikely that every customer would respond to such a survey. One of the alternatives is to monitor customers' sentiment throughout the chat session. Likely, chat sessions with the most positive sentiment dialogues or chats ending with positive sentiment dialogues are more likely to infer a positive experience and so on.

In recent years, fine-tuned pre-trained LMs such as BERT~\cite{Devlin2019BERTPO}, and RoBERTa~\cite{Liu2019RoBERTaAR}, which are trained on millions of documents in a self-supervised fashion to obtain general-purpose language representations, have achieved enormous success in many natural language processing (NLP) tasks. While there is lots of literature available performing sentiments analysis by fine-tuning these LMs, most of this literature focuses on movie/product/restaurant/service reviews or tweets due to the availability of labeled public datasets for such use cases. To best to our knowledge, there is no labeled data available for sentiment analysis for customer chat.

Despite their enormous success for various NLP tasks, pre-trained LMs still require excessive labeled data during fine-tuning stage. These labeled training data are expensive and time-consuming to create, often requiring person-months or years to assemble, clean, and debug, especially when domain expertise is required. On top of this, tasks often change and evolve in the real world. For example, labeling guidelines, granularities, or downstream use cases often change, necessitating re-labeling. For example, instead of classifying sentiments only as positive or negative, introduce a neutral category. For all these reasons, practitioners have increasingly been turning to weaker forms of supervision, such as heuristically generating training data with external knowledge bases, patterns/rules, or other classifiers. For example, for NLP in sentiment analysis, we can use rules 'terrible'$->$Negative (a keyword rule) and '* not recommend *'$->$ Negative (a pattern rule) to generate large amounts of weakly labeled data.

The central hypothesis of this paper is: For customer sentiment analysis in customer-agent chat, in absence of large-scale labeled training data, we can use weak labels for unlabeled data and then use these weakly labeled data to fine-tune LMs such as BERT, RoBERTa for sentiment classification using weak supervision. Weak labeling provides flexibility to add domain-specific rules for labeling. LMs fine-tuned using such weak supervision have an advantage for the domain-specific sentiment analysis compared to any generic off-the-shelf sentiment analysis solution. We can use customers' sentiment analysis during a chat session to understand their satisfaction with the chat service. Figure~\ref{fig:sentiment_classification} summarizes our training approach. We use five LFs to weakly label sentiments. Three weak classifiers Textblob~\cite{loria2018textblob}, AFINN~\cite{NielsenF2011New}, VADER~\cite{Hutto2014VADERAP} and two domain-specific lexicon-based LFs one each for positive and negative sentiment. We use Snorkel label model~\cite{ratner2018training}  to generate one final weak label from these 5 labels. We fine-tune the pre-trained RoBERTa model using weak supervision as per the approach mentioned in~\cite{yu2021finetuning}. Additionally, we get sentiment results on test data using google cloud NLP API for sentiment analysis and show how a fine-tuned model trained using weak supervision performs better for certain domain-specific sentiment classifications. We also perform an analysis on how customer sentiment relates to problem resolution during a chat. 

Rest of the paper is as follow: Section~\ref{sec:Related Work} describes related work, Section~\ref{sec:Data} describes data used in experiment, In  section~\ref{sec:Models} we describe models used in this experiment. Section~\ref{sec:Experiments} describes experiments details. Analysis is described ion section~\ref{sec:Analysis} and  conclusion is in section~\ref{sec:Conclusion}. 

\section{Related Work}
\label{sec:Related Work}
In this section, we summarize some prior work done for model training in absence of availability of large-scale training data. Snorkel~\cite{ranter2017snorkel} proposes a system that allows users to train models without hand labeling any training data. Users can write LFs that express arbitrary heuristics without worrying about correlation or accuracy. They expand work in~\cite{ratner2018training} and present a generative model to estimate true labels using weak labels as noisy observation. UST~\cite{mukherjee2020uncertaintyaware}  proposes LM fine-tuning mechanism using very few labeled data per class. It shows promising results by using only 20-30 samples per class on five text classification public datasets. This work is inspired by the self-training approach in which a teacher model is trained using few labeled data and used to pseudo-annotate task-specific unlabeled data. The original data is augmented with pseudo labeled data and used to train the student model. The student-teacher training is repeated until convergence. They use pre-trained LMs as a Teacher and Student models. Denoise~\cite{Ren2020denoising} addresses two issues of weak supervision: 1)Rule-based labels are noisy because heuristic rules/LFs are often too simple to capture rich contexts and complex semantics for texts, 2) Rule-based weak supervision does not cover a certain portion of available unlabeled data as some of the data isn't covered by any defined LF. It performs weighted aggregation of the predictions from multiple LFs using a soft attention mechanism, which generates higher-quality pseudo labels and makes the model less sensitive to the error in one single source. It also uses temporal ensembling, which aggregates historical pseudo labels and alleviates noise propagation. X-Class~\cite{wang2020xclass} proposes a fine-tuning mechanism with extremely weak supervision, where the training process can start with just a single word per class or only class names as supervision. It first performs class-oriented document representation estimation followed by document-class alignment through clustering using Gaussian Mixture Model (GMM). In the end, based on the posterior probability of a document to its assigned cluster, most confident documents are chosen for supervised training. \textbf{Co}ntrastive-regularized \textbf{S}elf-training for F\textbf{ine}-tuning (COSINE) ~\cite{yu2021finetuning} addresses the issue of overfitting of LM to noisy weak labels by introducing a contrastive self-training framework. This framework uses confidence-based sample reweighing and confidence regularization to make sure LM doesn't overfit noisy labels.

Among all reviewed work, COSINE has the best results for weak supervision. It shows accuracy for binary (positive, negative) sentiment classification using weak supervision close to accuracy using full supervision for IMDB~\cite{maas-EtAl:2011:ACL-HLT2011} and Yelp~\cite{meng2018yelp} sentiment classification datasets. However, it only uses lexicon-based rules as LFs and lexicon-based rules are focused on movie/restaurant reviews. Our approach differs from them as in addition to client-agent chat domain-specific lexicon-based rules, we use weak classifiers as LFs. We also use neutral class in addition to positive and negative.

\section{Data}
\label{sec:Data}
Our primary dataset for this work is an in-house dataset derived from actual customer-agent chats for customer support. Following are details about the dataset.  
Training Set: 320K unlabeled customer sentences from 10K chat sessions.
Dev Set: 610 manually labeled customer sentences from 37 chat sessions. 
Test Set: 621 manually labeled customer sentences from 41 chat sessions.

In addition to the in-house dataset, we share the result of the partial approach on the Stanford Sentiment Tree-bank (SST-3) dataset~\cite {NielsenF2011New}as well. Following are details about the secondary dataset:
Training Set: 8544 sentences.
Dev Set: 1101 sentences.
Test Set: 2210 sentences.

\section{Models}
\label{sec:Models}
\begin{figure*}[t]
  \centering
  \includegraphics[width=\textwidth]{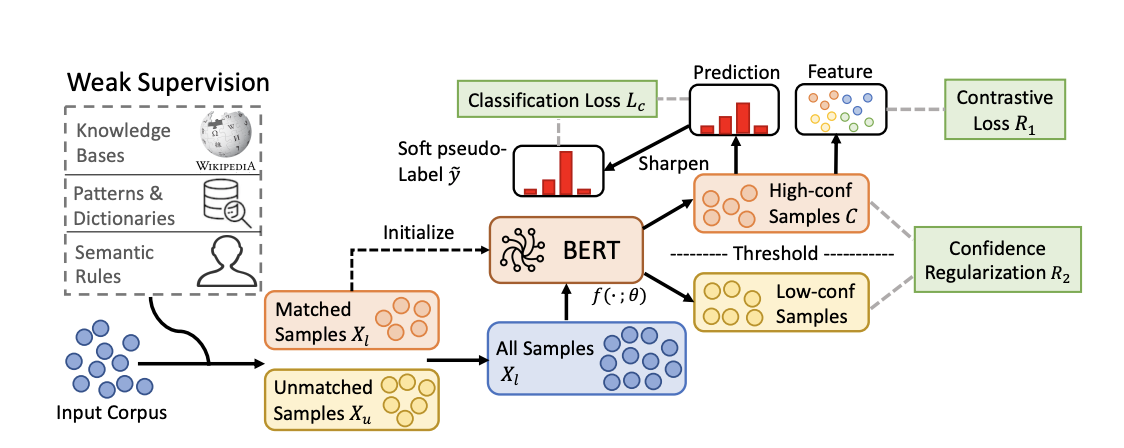}
  \caption{The framework of COSINE. We first fine-tune the pre-trained language model on weakly-labeled data with early stopping. Then, we conduct contrastive-regularized self-training to improve model generalization and reduce the label noise. During self-training, we calculate the confidence of the prediction and update the model with high confidence samples to reduce error propagation.}
  \label{fig:cosine_training}
\end{figure*}
We rely on the following modeling approaches for our experiments: weak classifiers, baseline model, LM fine-tuning

\subsection{Weak Classifiers}
\label{subsec:Weak Classifiers}
To create weak labels, we use three different weak sentiment classifiers as LFs:  Textblob~\cite{loria2018textblob}, AFINN~\cite{NielsenF2011New}, VADER~\cite{Hutto2014VADERAP}.

Textblob is a Python library for processing textual data. It provides a simple API for diving into common natural language processing (NLP) tasks such as part-of-speech tagging, noun phrase extraction, sentiment analysis, classification, translation, and more. It manages a dictionary for lexicons with polarity, subjectivity, and intensity score and applies different statistical rules as per the presence of certain words/phrases in a given sentence.

The AFINN lexicon is a list of English terms manually rated for valence with an integer between -5 (negative) and +5 (positive) by Finn Årup Nielsen between 2009 and 2011. The newest version has 2477 unique words, including 15 phrases. The final score for a given sentence is the sum of valance scores of different terms present in the sentence.

VADER (Valence Aware Dictionary for sEntiment Reasoning) is a simple rule-based model for general sentiment analysis. Using a combination of qualitative and quantitative methods, it first constructs and empirically validates a gold-standard list of lexical features (along with their associated sentiment intensity measures) which are specifically attuned to sentiment in microblog-like contexts. It then combine these lexical features with consideration for five general rules that embody grammatical and syntactical conventions for expressing and emphasizing sentiment intensity.

\subsection{Baseline Model}
\label{subsec:Baseline Model}
We use Snorkel label model~\cite{ratner2018training} to generate a single label from different LFs. This label model learns LFs' conditional probabilities of outputting the true (unobserved) label $Y, P( LF | Y )$. It re-weights and combines the output of different LFs and generates a single label. This approach computes the inverse generalized covariance matrix of the junction tree of a given LF dependency graph, and perform a matrix completion-style approach with respect to these empirical statistics. The result is an estimate of the conditional LF probabilities, $P(LF | Y)$, which are then set as the parameters of the label model used to re-weight and combine the labels output by the LFs. We use this model as a baseline model.

\subsection{LM Fine-tuning}
\label{subsec:LM Fine-tuning}
We fine-tune pre-trained RoBERTa~\cite{Liu2019RoBERTaAR} model with weak supervision using  contrastive-regularized self-training (COSINE) approach per ~\cite{yu2021finetuning}. In this approach, first, a pre-trained LM is initialized with a fraction of weakly labeled data. With this exercise, semantic and syntactic knowledge of the pre-trained LM is transferred to the new model. Then this new model is further trained using contrastive learning. Contrastive learning encourages data within the same class to have similar representations and keeps data in different classes separated. It enforces representations of samples from different classes to be more distinguishable, such that the classifier can make better decisions. It uses the following two steps to accomplish this objective: 1) Confidence-based sample reweighing. It gives higher weights to samples that have higher prediction confidence, and  2) Confidence Regularization. It encourages smoothness over model predictions, such that no prediction can be over-confident, and therefore reduces the influence of wrong pseudo-labels. Figure~\ref{fig:cosine_training} shows the training approach used for fine-tuning pre-trained LM as mentioned in~\cite{yu2021finetuning}.

\section{Experiments}
\label{sec:Experiments}

\subsection{Model Training}
\label{subsec:Model Training}
We use 5 LFs to weakly label data. Three LFs are based on weak classifiers as described in~\ref{subsec:Weak Classifiers}. We use polarity returned by Textblob for a given sentence to determine its sentiment. If the polarity is below -0.1 we label the sentence as negative, if it's above +0.1 we label it as positive else neutral. For AFINN we use final score of a sentence to determine sentiment. If the final score is below 0 we label the sentence as negative, if it's above 0 we label it as positive else neutral. We use the compound value of polarity score returned by VADER sentiment analyzer to determine sentiment. If the compound score is below -0.1 we label the sentence as negative, if it's above +0.1 we label it as positive else neutral. We choose threshold values for each classifier based on optimal results for the dev sets. 

In addition to the 3 weak classifiers, we use domain-specific lexicon-based LFs for positive and negative sentiment. The domain-specific lexicon has 65 negative terms an 15 positive terms. Here are some examples from domain-specific negative terms:~\textit{att, verizon} - because they are competitors,~\textit{call} -because it's a chat conversation,~\textit{manager}- because it means escalations. Here are a few examples from the positive terms:~\textit{sign off, goodbye} - because both these terms appear primarily when the customer happily ends the conversation. This lexicon is chosen manually based on the labeled dev set.

Snorkel label model as described in~\ref{subsec:Baseline Model} is used to generate a single label from 5 LFs. We use this label model as baseline model. We use RoBERTa as LM and fin-tune it using weak supervision COSINE framework~\ref{subsec:LM Fine-tuning}. Our hyper-parameter configuration for COSINE is summarized in Appendix~\ref{Hyper-parameters for RoBERTa fine-tuning using COSINE}.

We use google cloud NLP API for sentiment analysis to compare results of LM fine-tuning with an off-the-shelf solution. We use sentiment score returned by the api to determine sentiment of the solution. If the score is below -0.1 we label the sentence as negative, if it's above +0.1 we label it as positive else neutral.
\subsection{Metrics}
\label{subsec:Metrics}
We measure performance of algorithm using following two metrics:1)~macro F1 score. The majority of dialogues from customers tend to have negative sentiment for customer-agent chats, so we have an imbalanced representation for each sentiment. From an evaluation perspective, we'd like to give equal weightage to each class hence the macro F1 score is the right choice for that purpose. 2)~accuracy. Many prior works~\cite{yu2021finetuning, wang2020xclass}, for similar tasks, have presented results using accuracy. For consistency with prior works, we'll also evaluate the accuracy of the model.
\subsection{Results}
\label{subsec:Results}
Table~\ref{Model Performance} describes the performance of models on the in-house customer chat test set. Textblob has the lowest macro F1-score of 0.36 among all the weak classifiers while VADER has the highest macro score of 0.5. The snorkel label model, which uses 5 LFs, improves the macro F1-score to 0.52. RoBERTa fine-tuning using weak supervision through the COSINE framework achieves a macro F1-score of 0.56 and 0.65 after the init and final stages respectively. Google cloud NLP API for sentiment analysis has a macro F1-score of 0.69. As shown, even with weak supervision, we achieve accuracy close to that of google cloud NLP API for our dataset. Results for SST-3 are summarized in Appendix~\ref{sec:SST-3 Results}. For SST-3 only difference is we don't use domain-specific lexicon-based LFs.
\begin{table}
\centering
\begin{tabular}{l c c }
\hline \textbf{Model} & \textbf{Macro} & \textbf{Accuracy}  \\ 
&\textbf{F1-score} &\\
\hline
Textblob & 0.36 & 0.35  \\
AFINN & 0.46 & 0.46  \\
VADER & 0.50 & 0.49  \\
\hline
Baseline & 0.52 & 0.52 \\
\hline
RoBERTa fine-tuning &  &  \\
using COSINE & &  \\
Init & 0.56 & 0.57 \\
Final & 0.65 & 0.69 \\
\hline
Google cloud NLP API & 0.69 & 0.72\\
\hline
\end{tabular}
\caption{\label{Model Performance} Model performances on Internal Test Set}
\end{table}

\section{Analysis}
\label{sec:Analysis}
\begin{figure}[t]
  \centering
  \includegraphics[width=0.5\textwidth]{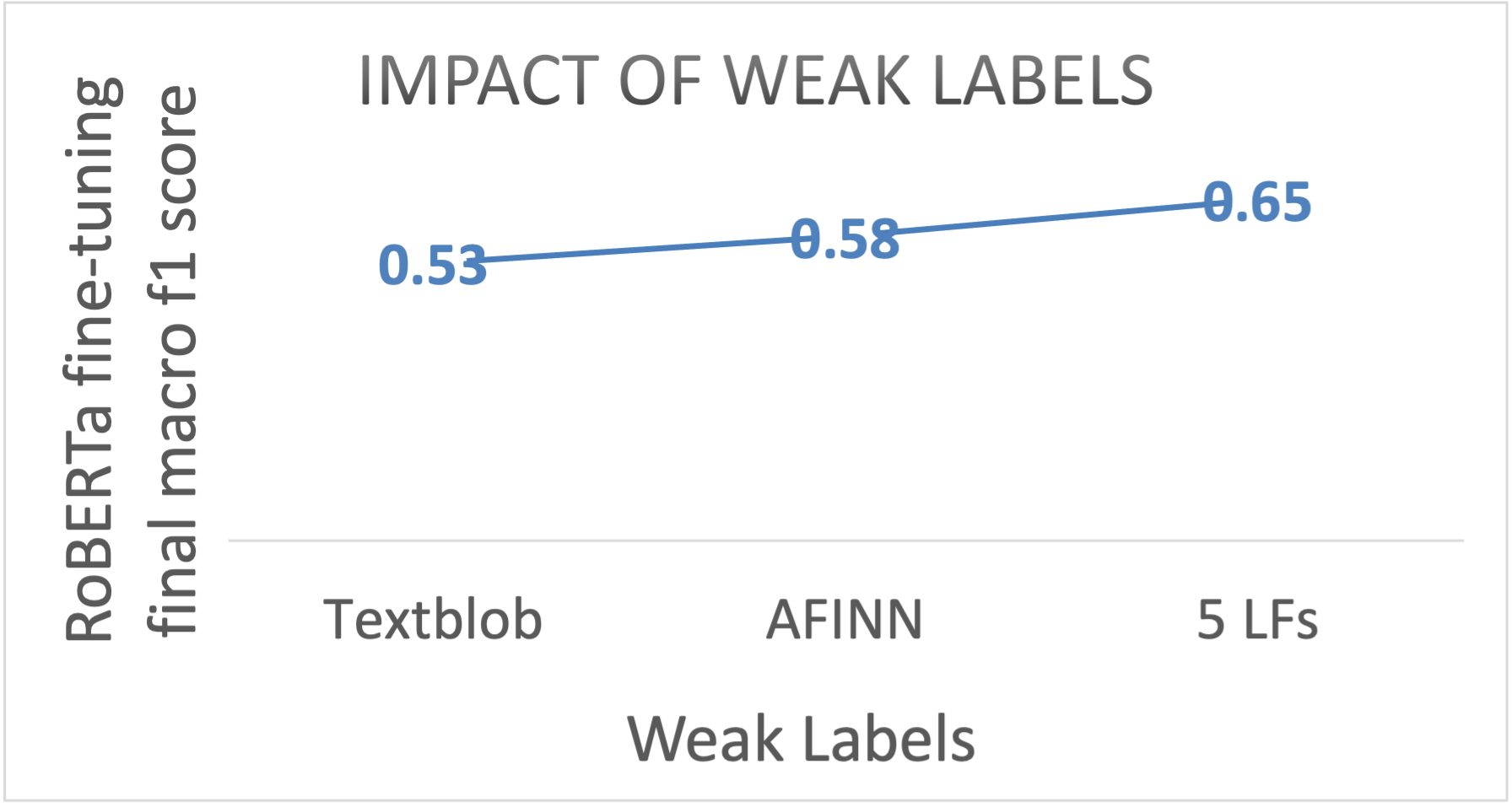}
  \caption{Impact of weak labels. More accurate weak labels lead to a more accurate fine-tuned LM.}
  \label{fig:weak_label_imapct}
\end{figure}
\begin{figure*}[t]
  \centering
  \includegraphics[width=\textwidth]{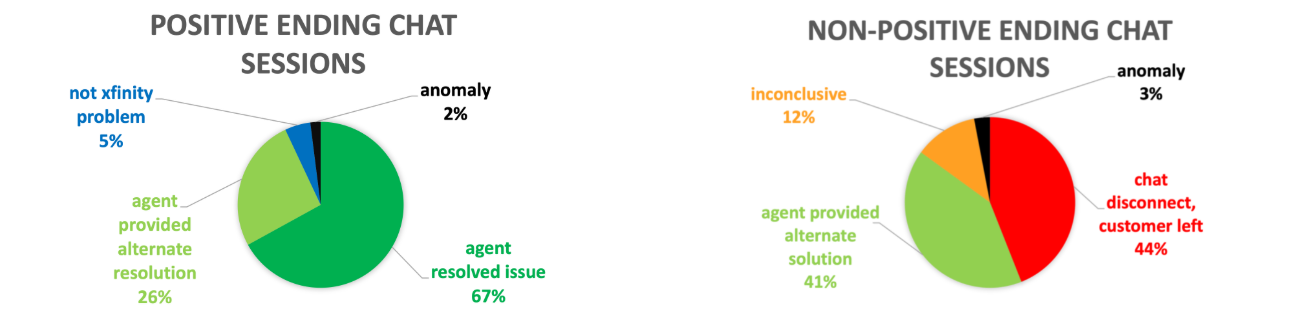}
  \caption{Sentiment and Problem Resolution. Our analysis on test data shows if an agent manages to resolve issue during a chat session, such chat ends with positive sentiment. If a customer leaves chat then such chat ends with non-positive sentiment. }
  \label{fig:sentiment_end}
\end{figure*}
\subsection{Impact of weak labels}
\label{subsec:Impact of weak labels}
We analyze the impact of weak labels on the final fine-tuning of LM. We compare results of three weak labels: labels created using 1)  only Textblob, 2) only AFINN, 3)  snorkel label model using 5 LFs. Figure~\ref{fig:weak_label_imapct} summarizes impact of these weak labels on final macro f1-score. As shown in Table~\ref{Model Performance}, Textblob, AFINN, and snorkel label model using 5 LFs have macro f1-score of 0.36, 0.46, and 0.52 respectively. After fine-tuning of RoBERTa using the COSINE framework, we get a final macro f1-score of 0.53, 0.58, and 0.65 for labels generated using Textblob, AFINN, and snorkel label model respectively. It indicates the more accurate weak labels lead to the better fine-tuning of LM.

\subsection{Comparison with Google Cloud NPL API}
\label{subsec:Comparison with Google Cloud NPL API}
\begin{table}
\centering
\begin{tabular}{ l }
\hline \textbf{domain-specific negative sentiment but}\\
\textbf{google cloud NLP API predicts positive}\\
\hline
ATT has Fiber and it’s only \$80.\\
thats insane !\\
\hline
 \textbf{domain-specific positive sentiment but}\\
\textbf{google cloud NLP API predicts negative}\\
\hline
good bye\\
 No, thank you, you don't have to book it.\\
\hline

\end{tabular}
\caption{\label{google comparison} Some examples where google cloud NLP APIs predict sentiment wrong but our RoBERTa model fine-tuned using weak supervision gets right. }
\end{table}
Table~\ref{google comparison} shows some examples where google cloud NLP API predicts the wrong sentiment for the domain-specific context but our fine-tuned RoBERTa model gets them right. For instance, in general sense \textit{ATT has Fiber and it’s only \$80} is a positive sentiment dialogue. But, for Comcast customer chat, someone referring to competitor ATT represents negative sentiment. Similarly, \textit{good bye} in general has negative sentiment as it relates to departure, but for a chat session, if a customer takes time to say goodbye, it's a positive thing. It shows that by using domain-specific lexicon-based LFs, we can tackle such issues, which won't be feasible with any off-the-shelf solution such as google cloud NLP API. We want to note that, as shown in Table~\ref{Model Performance}, overall, google cloud NLP API still performs better compared to RoBERTa fine-tuned with weak supervision, but it misses out on domain-specific challenges.

\subsection{Sentiment and Problem Resolution}
\label{subsec:Sentiment and Problem Resolution}
We also analyze how customer sentiment at the end of conversation relates to problem resolution. Result of this analysis is summarized in Figure~\ref{fig:sentiment_end}. If an agent manages to resolve a customer’s issue during a chat, such chat ends with positive sentiment. If a customer leaves a chat indicating unresolved issues, such chats end with a non-positive sentiment. But, if the agent provides an alternative solution such as call transfer,  technician appointment etc. during a chat, such chat can end with positive, neutral, or negative sentiment. 
\section{Conclusion}
\label{sec:Conclusion}

In absence of the availability of large-scale labeled data, we can use weak supervision to fine-tune LM to train a sentiment classifier to analyze customer sentiments during a chat session. We can improve the domain-specific result for such a classifier by using domain-specific lexicon-based LFs in addition to weak classifiers. Our result shows, a classifier trained using such weak labels can identify certain domain-specific sentiments better than any off-the-shelf solution such as google cloud NLP API. Moreover, our analysis shows problem resolution during chat results in chat ending with positive sentiment. If an agent doesn't resolve the issue during the chat but provide an alternate solution such as call transfer, technician appointment, etc., customers' sentiment varies in such situation. Every time a customer leaves a chat, such chats have non-positive sentiment at the end.

For future work, we would like to explore an automated way to identify domain-specific sentiment-related terms. We'd also like to verify the limitation of weak labels eg. if we use very accurate labels as weak labels, how it impacts LM fine-tuning. We also want to explore different LM fine-tuning techniques. We would like to analyze a correlation between sentiment and NPS and see whether sentiment can be used as a proxy for NPS.

\section{Acknowledgement}
\label{sec:Acknowledgement}
I thank Christopher Potts, Professor, and Chair, Department of Linguistics and Professor, by courtesy, Department of Computer Science, Stanford University, for mentoring me throughout different stages of this work. I thank Steve Haraguchi, Director of AI Programs, Stanford University, for facilitating XCS224U. I thank Powell Molleti, my course facilitator for XCS224U, for providing me feedback during different stages of this work. I thank Ferhan Ture, Director of AI for voice/NLP, Comcast Applied AI Research, for helping me finalize this work and providing feedback throughout the development of this work. I thank Hongcheng Wang, Director of AI for connected living and customer experience, and my manager, Comcast Applied AI Research, for accommodating this work. 

\bibliography{anthology,acl2020}

\begin{thebibliography}{13}
\expandafter\ifx\csname natexlab\endcsname\relax\def\natexlab#1{#1}\fi

\bibitem[{Devlin et~al.(2019)Devlin, Chang, Lee, and
  Toutanova}]{Devlin2019BERTPO}
Jacob Devlin, Ming-Wei Chang, Kenton Lee, and Kristina Toutanova. 2019.
\newblock Bert: Pre-training of deep bidirectional transformers for language
  understanding.
\newblock In \emph{NAACL}.

\bibitem[{Hutto and Gilbert(2014)}]{Hutto2014VADERAP}
Clayton~J. Hutto and Eric Gilbert. 2014.
\newblock Vader: A parsimonious rule-based model for sentiment analysis of
  social media text.
\newblock In \emph{ICWSM}.

\bibitem[{Liu et~al.(2019)Liu, Ott, Goyal, Du, Joshi, Chen, Levy, Lewis,
  Zettlemoyer, and Stoyanov}]{Liu2019RoBERTaAR}
Yinhan Liu, Myle Ott, Naman Goyal, Jingfei Du, Mandar Joshi, Danqi Chen, Omer
  Levy, Mike Lewis, Luke Zettlemoyer, and Veselin Stoyanov. 2019.
\newblock Roberta: A robustly optimized bert pretraining approach.
\newblock \emph{ArXiv}, abs/1907.11692.

\bibitem[{Loria(2018)}]{loria2018textblob}
Steven Loria. 2018.
\newblock textblob documentation.
\newblock \emph{Release 0.15}, 2.

\bibitem[{Maas et~al.(2011)Maas, Daly, Pham, Huang, Ng, and
  Potts}]{maas-EtAl:2011:ACL-HLT2011}
Andrew~L. Maas, Raymond~E. Daly, Peter~T. Pham, Dan Huang, Andrew~Y. Ng, and
  Christopher Potts. 2011.
\newblock \href {http://www.aclweb.org/anthology/P11-1015} {Learning word
  vectors for sentiment analysis}.
\newblock In \emph{Proceedings of the 49th Annual Meeting of the Association
  for Computational Linguistics: Human Language Technologies}, pages 142--150,
  Portland, Oregon, USA. Association for Computational Linguistics.

\bibitem[{Meng et~al.(2018)Meng, Shen, Zhang, and Han}]{meng2018yelp}
Yu~Meng, Jiaming Shen, Chao Zhang, and Jiawei Han. 2018.
\newblock \href {https://doi.org/10.1145/3269206.3271737} {Weakly-supervised
  neural text classification}.
\newblock \emph{Proceedings of the 27th ACM International Conference on
  Information and Knowledge Management}.

\bibitem[{Mukherjee and Awadallah(2020)}]{mukherjee2020uncertaintyaware}
Subhabrata Mukherjee and Ahmed~Hassan Awadallah. 2020.
\newblock \href {http://arxiv.org/abs/2006.15315} {Uncertainty-aware
  self-training for text classification with few labels}.

\bibitem[{Nielsen(2011)}]{NielsenF2011New}
Finn~{\AA}rup Nielsen. 2011.
\newblock \href {http://ceur-ws.org/Vol-718/paper_16.pdf} {A new {ANEW}:
  evaluation of a word list for sentiment analysis in microblogs}.
\newblock In \emph{Proceedings of the ESWC2011 Workshop on 'Making Sense of
  Microposts': Big things come in small packages}, volume 718 of \emph{CEUR
  Workshop Proceedings}, pages 93--98.

\bibitem[{Ratner et~al.(2017)Ratner, Bach, Ehrenberg, Fries, Wu, and
  Ré}]{ranter2017snorkel}
Alexander Ratner, Stephen~H. Bach, Henry Ehrenberg, Jason Fries, Sen Wu, and
  Christopher Ré. 2017.
\newblock \href {https://doi.org/10.14778/3157794.3157797} {Snorkel}.
\newblock \emph{Proceedings of the VLDB Endowment}, 11(3):269–282.

\bibitem[{Ratner et~al.(2018)Ratner, Hancock, Dunnmon, Sala, Pandey, and
  Ré}]{ratner2018training}
Alexander Ratner, Braden Hancock, Jared Dunnmon, Frederic Sala, Shreyash
  Pandey, and Christopher Ré. 2018.
\newblock \href {http://arxiv.org/abs/1810.02840} {Training complex models with
  multi-task weak supervision}.

\bibitem[{Ren et~al.(2020)Ren, Li, Su, Kartchner, Mitchell, and
  Zhang}]{Ren2020denoising}
Wendi Ren, Yinghao Li, Hanting Su, David Kartchner, Cassie Mitchell, and Chao
  Zhang. 2020.
\newblock \href {https://doi.org/10.18653/v1/2020.findings-emnlp.334}
  {Denoising multi-source weak supervision for neural text classification}.
\newblock \emph{Findings of the Association for Computational Linguistics:
  EMNLP 2020}.

\bibitem[{Wang et~al.(2020)Wang, Mekala, and Shang}]{wang2020xclass}
Zihan Wang, Dheeraj Mekala, and Jingbo Shang. 2020.
\newblock \href {http://arxiv.org/abs/2010.12794} {X-class: Text classification
  with extremely weak supervision}.

\bibitem[{Yu et~al.(2021)Yu, Zuo, Jiang, Ren, Zhao, and
  Zhang}]{yu2021finetuning}
Yue Yu, Simiao Zuo, Haoming Jiang, Wendi Ren, Tuo Zhao, and Chao Zhang. 2021.
\newblock \href {http://arxiv.org/abs/2010.07835} {Fine-tuning pre-trained
  language model with weak supervision: A contrastive-regularized self-training
  approach}.

\end{thebibliography}
\bibliographystyle{acl_natbib}

\appendix

\section{Hyper-parameters for RoBERTa fine-tuning using COSINE}
\label{Hyper-parameters for RoBERTa fine-tuning using COSINE}
Our hyper-parameters for RoBERTa fine-tuning using COSINE for both Customer chat and SST-3 datasets are summarized in Table~\ref{Hyper-Parameter COSINE}.
\begin{table}[!ht]
\begin{tabular}{c |c |c }
\hline \textbf{Hyper-parameter} & \textbf{Customer Chat} & \textbf{SST-3}  \\ 
\hline
Dropout Ratio & \multicolumn{2}{c}{0.1}  \\
\hline
Maximum Tokens & \multicolumn{2}{c}{128} \\
\hline
Batch Size & 256 & 32 \\
\hline
Weight Decay & $10^{-5}$ & $10^{-4}$\\
\hline
Learning Rate & \multicolumn{2}{c}{$10^{-5}$} \\
\hline
$T_1$ & 120 & 350\\
\hline
$T_2$ & 600 & 2500\\
\hline
$T_3$ & 120 & 250\\
\hline
$\xi$ & \multicolumn{2}{c}{0.6} \\
\hline 
$\lambda$  &\multicolumn{2}{c}{0.1}\\
\hline
\end{tabular}
\caption{\label{Hyper-Parameter COSINE} Hyper-parameter configurations for RoBERTa fine-tuning using COSINE}
\end{table}
\label{sec:Hyperparameter for RoBERTa fine-tuning using COSINE}

\section{SST-3 Results}
\label{sec:SST-3 Results}
Table~\ref{SST-3 Performance} shows the result of different classifiers on SST-3 test set. Here baseline snorkel label model only has 3 weak classifiers as LFs.
\begin{table}[!ht]
\begin{tabular}{l |c| c }
\hline \textbf{Model} & \textbf{Macro} & \textbf{Accuracy}  \\ 
&\textbf{F1-score} &\\
\hline
Textblob & 0.45 & 0.47  \\
AFINN & 0.46 & 0.5  \\
VADER & 0.45 & 0.49  \\
\hline
Baseline & 0.46 & 0.48 \\
\hline
RoBERTa fine-tuning &  &  \\
using COSINE & &  \\
Init & 0.52 & 0.55 \\
Final & 0.55 & 0.63 \\
\hline
Google cloud NLP API & 0.59 & 0.65\\
\hline
\end{tabular}
\caption{\label{SST-3 Performance} Model performances on SST-3 Test Set}
\end{table}

\end{document}